\pdfoutput=1
\documentclass[11pt]{article}

\usepackage{acl}
\usepackage{comment}
\usepackage{multirow}
\usepackage{times}
\usepackage{latexsym}
\usepackage{subcaption}
\usepackage{rotating}

\usepackage{xcolor}
\usepackage{amssymb}
\usepackage{geometry}
\usepackage[T1]{fontenc}
\DeclareUnicodeCharacter{1EBF}{\'{e}}   % ế
\DeclareUnicodeCharacter{1EA1}{\d{a}}   % ạ
\DeclareUnicodeCharacter{1EE7}{\u{u}}   % ủ
\DeclareUnicodeCharacter{1ED9}{\d{o}}   % ộ
\DeclareUnicodeCharacter{1EAB}{\~{a}}   % ẫ
\DeclareUnicodeCharacter{1EC3}{\~{e}}   % ễ
\DeclareUnicodeCharacter{1ED1}{\^{o}}   % ố
\DeclareUnicodeCharacter{1EEF}{\~{u}}   % ữ
\usepackage[utf8]{inputenc}
\usepackage{graphicx}
\usepackage{microtype}

\usepackage{color}
\usepackage{booktabs, tabularx}
\newcolumntype{C}{>{\centering\arraybackslash}X}
\usepackage{adjustbox}
\usepackage[T1]{fontenc}
\usepackage{booktabs}
\usepackage{multirow}
\usepackage[utf8]{inputenc}

\usepackage{microtype}

\usepackage{inconsolata}
\usepackage{graphicx}
\usepackage{graphicx,subcaption,booktabs,float}
\usepackage[utf8]{inputenc}
\usepackage[T1]{fontenc}
\usepackage{microtype}
\usepackage{amsmath}
\usepackage{graphicx}
\usepackage{multirow}
\usepackage{times}
\usepackage{latexsym}
\usepackage{subcaption}
\usepackage{rotating}
\usepackage{xcolor}
\usepackage{amssymb}
\usepackage{pgfplots}
\usepackage{pgfplotstable}

\title{Exploring Data and Parameter Efficient Strategies for Arabic Dialect Identifications}
\author{Vani Kanjirangat \\
  SUPSI, IDSIA, Switzerland \\
  \texttt{vanik@idsia.ch} 
  \\\And
  Ljiljana Dolamic \\
  armasuisse S+T, Switzerland  \\
  \texttt{Ljiljana.Dolamic@armasuisse.ch} 
  \\\And
  Fabio Rinaldi \\
  SUPSI, IDSIA, Switzerland \\
  \texttt{fabio.rinaldi@idsia.ch} 
}

\begin{document}
\maketitle
%https://arabicnlp2025.sigarab.org/
%June 22, 2025: Abstract submission for conference papers due date
%June 29, 2025: Conference paper due date

\begin{abstract}
This paper discusses our exploration of different data-efficient and parameter-efficient approaches to Arabic Dialect Identification (ADI). In particular, we investigate various soft-prompting strategies, including prefix-tuning, prompt-tuning, P-tuning, and P-tuning V2, as well as LoRA reparameterizations. For the data-efficient strategy, we analyze hard prompting with zero-shot and few-shot inferences to analyze the dialect identification capabilities of Large Language Models (LLMs). 
%The experimental evaluations were conducted mainly on Nuanced Arabic Dialect Identification (NADI) datasets. 
% We also comprehensively compare different Arabic models on the NADI datasets 2021-2023. In zero-shot settings, we also evaluated the LLMs on multi-label dialect classifications. 
For the parameter-efficient PEFT approaches, we conducted our experiments using Arabic-specific encoder models on several major datasets. We also analyzed the n-shot inferences on open-source decoder-only models, a general multilingual model (Phi-3.5), and an Arabic-specific one(SILMA). We observed that the LLMs generally struggle to differentiate the dialectal nuances in the few-shot or zero-shot setups. The soft-prompted encoder variants perform better, while the LoRA-based fine-tuned models perform best, even surpassing full fine-tuning.  
 \end{abstract}

\section{Introduction}

The task of Dialect identification (DI) focuses on classifying the given input utterance to a specific dialect class. Dialectal cues can be quite nuanced, with a lot of fine-grained overlaps ~\citep{zampieri2017findings,zampieri2018language}. In recent years, hard prompting has emerged as a simple yet effective and data-efficient approach for leveraging large language models (LLMs) in various natural language processing (NLP) tasks \cite{brown2020language}. Due to their extensive pre-training and advanced reasoning capabilities, LLMs enable zero-shot and few-shot inference, offering data-efficient solutions across a wide range of NLP tasks. While these models have demonstrated impressive performance, much focus has been on English-centric benchmarks \cite{ lai2023chatgpt}. In the context of dialect identifications (DI), studies exploring zero-shot and few-shot performance have primarily focused on open-source models, such as GPT and Gemini \cite{khondaker2023gptaraeval, bang2023multitask, rane2024gemini}. \\
Supervised fine-tuning allows encoder-based models to acquire domain-specific knowledge in the context of transfer learning \cite{pan2020transfer}. Applying this approach to LLMs introduces challenges such as pretrain-finetune discrepancy \cite{yang2019xlnet}, inherited pretraining biases, and the high computational cost of fine-tuning. To mitigate these issues, Parameter-Efficient Fine-Tuning (PEFT) techniques \cite{lialin2023scaling} have been introduced, which aim to achieve task adaptability with minimal parameter updates.\\
In this study, we focus on evaluating data-efficient and parameter-efficient strategies by selecting Arabic as the primary language and Arabic Dialect Identification (ADI) as the target task. Arabic is a language spoken by a large community of approximately 400 million people, which is widely distributed across various countries and regions.
The regional Arabic dialects differ from the Modern Standard Arabic (MSA) in lexical, syntactic, and phonetic aspects (MSA: the official language in many Arabic-speaking countries)\citep{zaidan2014arabic}.\\ 
% Some of the most popular datasets in ADI include: The ADI VarDial dataset~\citep{zampieri2017findings,zampieri2018language, malmasi2016discriminating,ali2016automatic}, Arabic Online Commentary (AOC)\citep{zaidan2011arabic}, the Multi Arabic Dialect Applications and Resources (MADAR) corpus \citep{bouamor2019madar}, etc. The NADI shared task started in 2020 and presented continued efforts in the ADI, including country-wise, province-wise, and region-wise dialect identification tasks \cite{abdul-mageed-etal-2020-nadi,abdul-mageed-etal-2021-nadi,abdul-mageed-etal-2022-nadi,abdul-mageed-etal-2023-nadi}. 
We define the research goals for the proposed work as: \emph{Firstly}, to analyze the data-efficient capabilities of LLMs in zero-shot and few-shot settings for the Arabic Dialect Identification (ADI) task. \emph{Secondly}, to evaluate and compare parameter-efficient fine-tuning methods, specifically LoRA and soft prompting, in the context of ADI tasks, and \emph{thirdly}, to perform a comparative analysis of prompting and fine-tuning strategies across multiple ADI datasets using Arabic-specific LLMs.
\section{Methods}
\vspace{-1mm}
In this section, we describe the approaches in detail: data-efficient and Parameter-efficient approaches.
\section{Data-efficient approaches}\label{app:prompt}
Although manual engineering of the prompts can be cumbersome, it remains practical and efficient in many applications. Prompting has emerged as a practical approach to infer LLMS without needing full fine-tuning \cite{brown2020language}. To analyze the dialect classification capabilities of LLMs, we focused on using the zero-shot (ZS) and few-shot (FS) inference strategies with LLMs.

We experimented with different ZS prompt variations, such as \emph{vanilla prompt}, \emph{chain-of-thought (CoT)} inspired prompting, and \emph{binary prompting}. For \emph{few-shot}, a general approach was utilized, where some sample input-output pairs from the training set were used to establish the few-shot examples. Further, as a strategy inspired by  \emph{Clue And Reasoning Prompting (CARP)} \cite{sun2023text} was also employed. CARP adopts a progressive reasoning strategy by first prompting the LLM to extract superficial clues (e.g., keywords, tones, semantic relations, references, etc), In our case, we prompted ChatGPT \cite{openai2023chatgpt} to get the specific dialectal vocabularies for each dialect and further used them as clues in the prompt. The templates of different prompting strategies are shown in Figures \ref{fig:prompt}, \ref{fig:prompt2}, and \ref{fig:prompt3} in Appendix \ref{app:prompts}.

\subsection{Parameter-Efficient Fine-Tuning (PEFT)}
PEFT includes a set of approaches that enable the efficient adaptation of LLMs in terms of memory and computational performance \cite{lialin2023scaling}. In this paper, we experimented with two variants of PEFT: \emph{reparameterization-based} and \emph{soft-prompting} methods.
% 1. reparameterization-based methods reduce the number of trainable parameters by utilizing low-rank representations, thereby reducing computational complexity; 2. soft-prompting, in contrast to conventional hard prompts, utilizes virtual tokens that are continuous and trainable.

\paragraph{Low-Rank Adaptation (LoRA)} LoRA \cite{hu2022lora} is a reparameterization method designed to adapt large pre-trained models with minimal additional parameters. Instead of updating the full weight matrices during training, LoRA freezes the original model weights and injects trainable low-rank matrices into specific layers, which significantly reduces memory and computational costs.
\paragraph{Prefix-Tuning}
Prefix-tuning \cite{li2021prefix} is a soft-prompt approach where a small set of learnable vectors — prefixes — are prepended to the input at each layer of a pre-trained language model (PLM). These prefixes can be interpreted as a sequence of virtual tokens that condition the model's internal representations without altering its original parameters. 
% By keeping the PLM frozen and optimizing only the prefix vectors ), prefix-tuning can achieve performance comparable to full fine-tuning. The goal is to learn a task-specific context that effectively steers the model's output toward solving a particular downstream task, making this approach both computationally efficient and highly modular.
\paragraph{Prompt-Tuning}
This soft-prompting \citep{lester2021power} involves using a PLM without any parameter updates and relies on natural language templates to guide the model's behavior. It introduces soft prompts, which are appended to the input embeddings. These soft prompts are optimized via backpropagation, while keeping the rest of the pre-trained model frozen. 
% For instance, in a sentiment analysis task, a sample like “Amazing movie!” might be paired with a prompt such as “This movie is [MASK]”, where the model is asked to predict whether the masked token should be “good” or “bad.” 
% This method requires no fine-tuning and stores only a single copy of the model’s parameters. However, discrete prompting, which uses fixed, manually designed prompts, can often yield suboptimal performance compared to methods that adapt model parameters. To address this, 

% This approach was introduced by , who demonstrated that tuning only these continuous prompts can improve performance on downstream tasks
% . However, despite its efficiency, prompt tuning tends to underperform full fine-tuning when applied to smaller models, particularly those with fewer than 10 billion parameters.

\paragraph{P-Tuning}
P-tuning \cite{liu2024gpt} is a soft-prompt designed to enhance the performance of language models like GPTs on Natural Language Understanding (NLU) tasks, where traditional fine-tuning often falls short. Unlike full fine-tuning, which updates all model parameters, P-tuning introduces trainable soft prompts that are prepended to the input. These embeddings are optimized to guide the model's behavior for specific downstream tasks, while the original parameters remain frozen. A distinctive feature of P-tuning is its use of a Long Short-Term Memory (LSTM) network to generate the soft prompts. The LSTM enables the model to capture sequential dependencies within the prompts, allowing more flexible and expressive task conditioning than static embeddings. 

% his dynamic prompt generation improves the alignment between the prompts and the structure of the input data, which is especially important for autoregressive models like GPTs.

% By combining soft prompts with LSTM-based generation, P-tuning can achieve performance on par with or better than full fine-tuning for many NLU tasks, while training far fewer parameters. It also helps overcome limitations of discrete prompts, such as brittleness and poor generalization, by learning continuous, differentiable representations that adapt better to the task.
\paragraph{P-TuningV2}
P-Tuning v2 \cite{liu2022p} extends the original P-Tuning approach and is designed to improve performance in both text generation and knowledge probing tasks. Functionally, it can be viewed as an application of Prefix-Tuning to encoder-based models, such as BERT. While earlier methods, such as prompt tuning, appended continuous prompts only at the input layer, P-Tuning v2 introduces deep prompt tuning, where trainable continuous prompts are inserted at every transformer layer. This significantly increases the capacity and expressiveness of the prompts, allowing the model to better capture task-specific nuances without modifying the underlying pre-trained parameters.

% One of the core insights behind P-Tuning v2 is that prompts injected into deeper layers have a more direct and impactful influence on model predictions compared to those added only at the input. This deep prompting strategy helps close the performance gap between prompt-based methods and full fine-tuning, especially on harder tasks and smaller models. While earlier research (Lester et al., 2021) showed that prompt tuning approaches full fine-tuning performance when scaling to very large models (10 B+ parameters), it often underperforms in medium-sized models (100M–1B), which are more commonly used. P-Tuning v2 addresses this limitation by increasing the number of trainable parameters—still in a parameter-efficient manner—ranging from 0.01\% to as much as 0.1–3\% of the full model, depending on the task. By integrating prompts throughout the transformer layers, P-Tuning v2 achieves more universal effectiveness across model sizes, making it a robust and scalable alternative to full fine-tuning.
\section{Datasets and Models}
\paragraph{Data:} In this work, we utilize different ADI datasets that cover a wide range of Arabic dialects, with different complexity levels. This includes Vardial ADI, Arabic Online Commentary (AOC), and NADI datasets. The VarDial ADI \cite{zampieri2017findings}, focused on five classes - \emph{MSA, Egyptian, Gulf, Levantine, Moroccan, North-African}. AOC \cite{zaidan2011arabic} covers \emph{MSA} and the dialectal varieties - \emph{ Egyptian, Gulf, Levantine, Moroccan}. NADI datasets, available since 2020 \cite{muhammad2020nadi}, were built upon and extended the MADAR dataset \cite{bouamor2019madar} by introducing a fine-grained, sub-country level dialect identification task. 
% While MADAR provides dialect labels at both city and country levels, it primarily targets the Twitter and travel domains. 
% In NADI 2021 \cite{abdul-mageed-etal-2021-nadi}, we introduced new datasets that distinguish between Modern Standard Arabic (MSA) and Dialectal Arabic (DA) based on geographical origin, covering 100 provinces across 21 Arab countries in the Twitter domain. 
% The VarDial ADI \cite{zampieri2017findings}, focused on five classes, viz., Modern Standard Arabic (MSA), Egyptian
% (EGY), Gulf (GLF), Levantine (LAV), Moroccan (MOR), and North-African (NOR). AOC \cite{zaidan2011arabic} covers MSA and the dialectal varieties, viz., Egyptian (EGY), Gulf (GLF), Levantine (LEV), and Moroccan (MOR).
NADI-2022 \cite{abdul-mageed-etal-2022-nadi} includes approximately 20,000 tweets across 18 dialects. The data statistics of each dataset are presented in Table \ref{tab:data}. Figure \ref{fig:country} in Appendix \ref{app:data} shows the label distribution of NADI datasets, showing that datasets from 2020 to 2022 are quite unbalanced. In contrast, the NADI-2023 dataset provides 18 dialects with a more balanced distribution \cite{abdul-mageed-etal-2023-nadi}.
% Additionally, Figure \ref{fig:data1} presents the overall dataset statistics and average sentence lengths across the NADI datasets.
\begin{table}
\centering
%\small
\begin{tabular}{lrrrr}
\hline
{} & \textbf{NADI} & \textbf{NADI} & \textbf{VarDial} & \textbf{AOC}  \\ 
{} & \textbf{-2022} & \textbf{-2023} & \textbf{-ADI} & \textbf{}  \\  
\hline
Train & 20398 &  18000& 21001 & 86541  \\
Dev  & 4871 &  1800& 1566 & 10820  \\ 
Test & 4871 & - & 1492 & 10812 \\
\hline
\end{tabular}
%}
\caption{\label{tab:data}The size of datasets expressed as the number of utterances. }
\end{table}

\begin{table*}[h]
\centering
\begin{tabular}{lcccc}
\hline
\textbf{Models} & \textbf{NADI 2022} & \textbf{NADI 2023} & \textbf{Vardial ADI} & \textbf{AOC} \\
\hline
AraBERT V.02        & 0.25 & 0.71  & 0.35 & 0.73 \\
AraBERT Twitter     & 0.29 & 0.79  & 0.39 & 0.73 \\
CamelBERT           & 0.24 & 0.73  & 0.35 & 0.72 \\
MultidialectBERT    & 0.27 & 0.73  & \textbf{0.41} & 0.71 \\
MARBERTV2           & \textbf{0.30}  & \textbf{0.84}  &   0.40    &  \textbf{0.79}     \\
\hline
\end{tabular}
\caption{Model performance across various Arabic dialect datasets using full fine-tuning}
\label{tab:result1}
\end{table*}
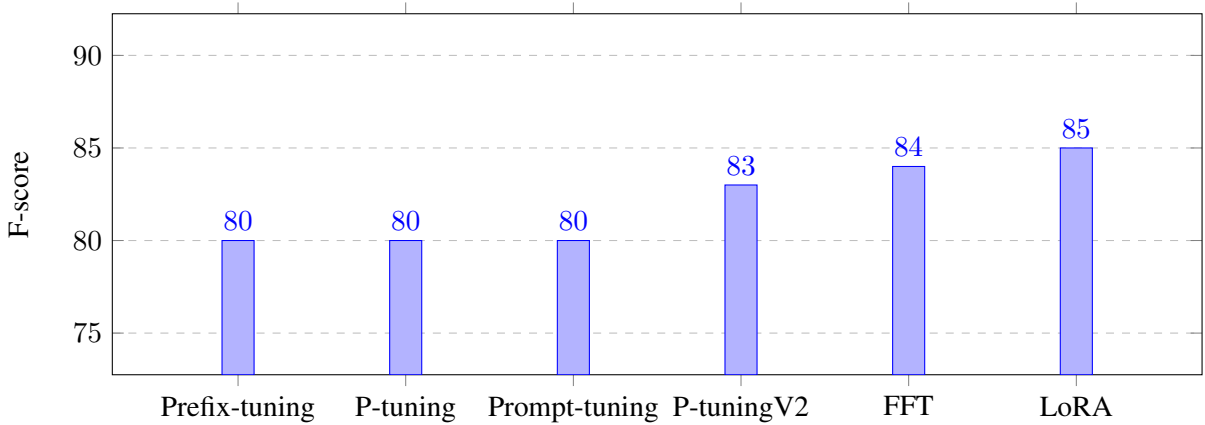
\begin{figure*}[h]
\centering
\begin{tikzpicture}
\begin{axis}[
    ybar,
    bar width=12pt,
    enlargelimits=0.15,
    ymin=75, ymax=90,
    ylabel={F-score},
    symbolic x coords={Prefix-tuning, P-tuning, Prompt-tuning, P-tuningV2, FFT, LoRA},
    xtick=data,
    nodes near coords,
    nodes near coords align={vertical},
    width=1.0\textwidth,
    height=0.4\textwidth,
    ymajorgrids=true,
    grid style=dashed
]
\addplot coordinates {
    (Prefix-tuning,80) 
    (P-tuning,80) 
    (Prompt-tuning,80) 
    (P-tuningV2,83)
    (LoRA,85)
    (FFT,84) 
    
};
\end{axis}
\end{tikzpicture}
\caption{Comparison of F-scores (\%) of various PEFT methods with Full Fine Tuning (in NADI-2023 ADI dataset)}
\label{fig:results2}
\end{figure*}
\vspace{-5mm}
\paragraph{Models:} We compare the following models under full fine-tuning (FFT)- \emph{AraBERT, AraBERT Twitter, CamelBERT, MultiDialectBERT, MARBERTv2}.
AraBERT \cite{antoun2020arabert} is an Arabic PLM based on BERT, trained with OSCAR unshuffled and filtered, Arabic Wikipedia dump, the 1.5B words Arabic Corpus, the OSIAN Corpus, and Assafir news articles. AraBERT Twitter was trained by continuing the pre-training using the MLM task on ~60M Arabic tweets (filtered from a collection of 100M). CamelBERT \cite{inoue2021interplay} is a collection of BERT models pre-trained on Arabic texts with different sizes and variants -  pre-trained language models for MSA, dialectal Arabic (DA), classical Arabic (CA), and a model pre-trained on a combination of the three. MultiDialectBERT \cite{talafha2020multi} is initialized with the model weights using Arabic-BERT and trained on 10M Arabic tweets from the unlabeled data of the NADI shared task. MARBERT \cite{mageed2021arbert} is pre-trained on very large and diverse datasets to facilitate transfer learning on MSA as well as Arabic dialect, along with a large Twitter dataset. It randomly samples 1B Arabic tweets from an extensive in-house dataset of about 6B tweets.

\section{Experimental Settings, Results \& Analysis}\label{sec:results}

% AraBERT- OSCAR unshuffled and filtered. Arabic Wikipedia dump from 2020/09/01. The 1.5B words Arabic Corpus
% The OSIAN Corpus (Arabic news articles), Assafir news articles
% Multidialect BERT-
% Based on araBERT (https://github.com/alisafaya/Arabic-BERT)
% Then trained  on 10M arabic tweets from the unlabled data of NADI task.

\subsection{Parameter-efficient approach results}
Table \ref{tab:result1} reports the FFT results across different ADI datasets on various Arabic-specific encoder models. After hyperparameter optimization, we used a dropout rate of 0.3, a learning rate (lr) of 1e-5, a batch size of 8, and 5 epochs. It can be observed that there is a wide discrepancy between the NADI-2022 and 2023 performances, where the latter was a balanced dataset. 

In Figure \ref{fig:results2}, we report the results with PEFT-based approaches with the MARBERTv2 model, since this model presented the best performances across various ADI datasets. For evaluation, we used NADI-2023, considering its complexity due to the inclusion of 18 dialects, while maintaining balanced distributions. We use LoRA with r=8 in parameterizations while keeping the other hyperparameters the same as FFT. In soft-prompting approaches, we compare the different techniques. For P-tuning, we used \emph{SEQ\_CLS} task, with lr of 1e-3, a weight\_decay of 0.01, and batch\_size of 8. We used a similar configuration for prompt-tuning and prefix-tuning with 20 virtual\_tokens, token\_dim of 768, num\_transformer\_submodules of 1, 12 attention\_heads, and 12 layers.  For these soft-prompting experiments, we used the corresponding HuggingFace wrappers. For the P-tuningV2, we adapted \citep{liu2022p}\footnote{\url{https://github.com/THUDM/P-tuning-v2}}.

It can be observed that all soft-prompting approaches performed similarly except P-tuningV2, presenting a F-score of 83\%, improving by 3 points. The best performance was achieved with LoRA, which outperformed even full fine-tuning by 1 point. 

\subsection{Data-efficient approach results}
One of the main challenges in ZS and FS inferences is to constrain the generation of LLMs to the predictions or classes we intended. Open-source LLMs are less straightforward than closed-source conversational models, such as ChatGPT or Gemini. For constraining the outputs properly, we relied on the library Skorch \footnote{\url{https://skorch.readthedocs.io/en/stable/}}. Skorch ensures that the model always predicts the expected labels by intercepting the model predictions (the logits) and forcing them to be one of the labels. We experimented with two open-source multilingual LLMs - \emph{Phi-3.5-mini} and \emph{SILMA}. SILMA is an Arabic-specific LLM with 9B parameters, based on the Gemma-7B model, and is essentially multilingual. We performed the inferences on the NADI-2023 dataset.

With the Phi-3.5-mini, we achieved a zero-shot (ZS) F-score of only 8\%, with a significant bias toward the Egyptian dialect. The situation did not improve with the Arabic-specific SILMA model, which showed a strong bias towards Saudi Arabian dialect. Removing the biased label did not resolve the issue, as the model consistently shifted its bias to a new label at each phase. We also attempted simple binary prompting (Figure \ref{fig:prompt2}) to reduce the complexity of the prompt. This resulted in a slight improvement, but not as much as expected. We also analyzed some samples using ChatGPT and Gemini, where the trend again appears to be biased towards Egyptian and Saudi Arabian dialects. The CARP-inspired approach did not yield the intended results. Instead, the clues seemed to act as noise. We could argue that the dialectal features generated by ChatGPT may not be suitable, and we may need to rely on manual curations, which could be expensive. Our experiments on Arabic dialects (NADI-2023) showed that LLMs' zero-shot or few-shot ability for dialectal discrimination is quite limited.  
%Further, we performed experiments by permutation of label orders to check whether the label order affects the predictions. This seemed to no effect, which means the order of label doesn't affected the LLM performances.

% For the GDI dataset, the ZS performance was 20\% with the Phi model, while the performance dropped to 10\% with the Llama model. In the ILI dataset, the ZS performance with the Llama model was only 9\%.
% In multi-label settings, we evaluated the French (FR) dataset with 200 samples; we obtained a macro-average F-score of 28\% and a weighted average of 45\%. The correctly predicted labels were predominantly FR-FR and FR-BE.

\textbf{Observations:} Model prediction may depend heavily on the pre-training data, such as the dialectal variety it has seen, without any specific understanding of the dialectal categorization features. With few shots, the samples may not be as indicative for this specific task as for other NLP tasks, such as sentiment analysis. In the specific use case of dialect classifications, the labels by themselves do not constitute any semantic meaning.

\section{Conclusion}\label{sec:conc}
In this paper, we present the analysis and comparison of the data-efficient prompting strategies and parameter-efficient fine-tuning approaches in the Arabic dialect identification task. We observed that the performance varies across the dialectal datasets based on the complexity and granularity of the dialectal classes. The performance across various Arabic-specific encoder variants shows that the PEFT approaches can be quite effective in these tasks. At the same time, the prompting strategies with LLMs reveal that regardless of the prompt variations, LLMs struggle to understand the nuanced dialectal cues. The challenge increases with the hard classification task, since dialectal varieties can often overlap, and sometimes these can be fine-lined.
\bibliography{custom}

\begin{thebibliography}{27}
\providecommand{\natexlab}[1]{#1}

\bibitem[{Abdul-Mageed et~al.(2024)Abdul-Mageed, Keleg, Elmadany, Zhang, Hamed, Magdy, Bouamor, and Habash}]{abdul-mageed-etal-2023-nadi}
Muhammad Abdul-Mageed, Amr Keleg, AbdelRahim Elmadany, Chiyu Zhang, Injy Hamed, Walid Magdy, Houda Bouamor, and Nizar Habash. 2024.
\newblock {NADI 2024: The Fifth Nuanced Arabic Dialect Identification Shared Task}.
\newblock In \emph{Proceedings of The Second Arabic Natural Language Processing Conference (ArabicNLP 2024)}.

\bibitem[{Abdul-Mageed et~al.(2022)Abdul-Mageed, Zhang, Elmadany, Bouamor, and Habash}]{abdul-mageed-etal-2022-nadi}
Muhammad Abdul-Mageed, Chiyu Zhang, AbdelRahim Elmadany, Houda Bouamor, and Nizar Habash. 2022.
\newblock \href {https://aclanthology.org/2022.wanlp-1.9} {{NADI} 2022: The third nuanced {A}rabic dialect identification shared task}.
\newblock In \emph{Proceedings of the The Seventh Arabic Natural Language Processing Workshop (WANLP)}, pages 85--97, Abu Dhabi, United Arab Emirates (Hybrid). Association for Computational Linguistics.

\bibitem[{Antoun et~al.()Antoun, Baly, and Hajj}]{antoun2020arabert}
Wissam Antoun, Fady Baly, and Hazem Hajj.
\newblock Arabert: Transformer-based model for arabic language understanding.
\newblock In \emph{LREC 2020 Workshop Language Resources and Evaluation Conference 11--16 May 2020}, page~9.

\bibitem[{Bang et~al.(2023)Bang, Cahyawijaya, Lee, Dai, Su, Wilie, Lovenia, Ji, Yu, Chung et~al.}]{bang2023multitask}
Yejin Bang, Samuel Cahyawijaya, Nayeon Lee, Wenliang Dai, Dan Su, Bryan Wilie, Holy Lovenia, Ziwei Ji, Tiezheng Yu, Willy Chung, et~al. 2023.
\newblock A multitask, multilingual, multimodal evaluation of chatgpt on reasoning, hallucination, and interactivity.
\newblock In \emph{Proceedings of the 13th International Joint Conference on Natural Language Processing and the 3rd Conference of the Asia-Pacific Chapter of the Association for Computational Linguistics (Volume 1: Long Papers)}, pages 675--718.

\bibitem[{Bouamor et~al.(2019)Bouamor, Hassan, and Habash}]{bouamor2019madar}
Houda Bouamor, Sabit Hassan, and Nizar Habash. 2019.
\newblock The madar shared task on arabic fine-grained dialect identification.
\newblock In \emph{Proceedings of the Fourth Arabic Natural Language Processing Workshop}, pages 199--207.

\bibitem[{Brown et~al.(2020)Brown, Mann, Ryder, Subbiah, Kaplan, Dhariwal, Neelakantan, Shyam, Sastry, Askell et~al.}]{brown2020language}
Tom Brown, Benjamin Mann, Nick Ryder, Melanie Subbiah, Jared~D Kaplan, Prafulla Dhariwal, Arvind Neelakantan, Pranav Shyam, Girish Sastry, Amanda Askell, et~al. 2020.
\newblock Language models are few-shot learners.
\newblock \emph{Advances in neural information processing systems}, 33:1877--1901.

\bibitem[{Hu et~al.(2022)Hu, Shen, Wallis, Allen-Zhu, Li, Wang, Wang, Chen et~al.}]{hu2022lora}
Edward~J Hu, Yelong Shen, Phillip Wallis, Zeyuan Allen-Zhu, Yuanzhi Li, Shean Wang, Lu~Wang, Weizhu Chen, et~al. 2022.
\newblock Lora: Low-rank adaptation of large language models.
\newblock \emph{ICLR}, 1(2):3.

\bibitem[{Inoue et~al.(2021)Inoue, Alhafni, Baimukan, Bouamor, and Habash}]{inoue2021interplay}
Go~Inoue, Bashar Alhafni, Nurpeiis Baimukan, Houda Bouamor, and Nizar Habash. 2021.
\newblock The interplay of variant, size, and task type in arabic pre-trained language models.
\newblock In \emph{Proceedings of the Sixth Arabic Natural Language Processing Workshop}, pages 92--104.

\bibitem[{Khondaker et~al.(2023)Khondaker, Waheed, Abdul-Mageed et~al.}]{khondaker2023gptaraeval}
Md~Tawkat~Islam Khondaker, Abdul Waheed, Muhammad Abdul-Mageed, et~al. 2023.
\newblock Gptaraeval: A comprehensive evaluation of chatgpt on arabic nlp.
\newblock In \emph{Proceedings of the 2023 Conference on Empirical Methods in Natural Language Processing}, pages 220--247.

\bibitem[{Lai et~al.(2023)Lai, Ngo, Veyseh, Mẫn, Dernoncourt, Bui, and Nguyen}]{lai2023chatgpt}
Viet Lai, Nghia Ngo, Amir Pouran~Ben Veyseh, Hiếu Mẫn, Franck Dernoncourt, Trung Bui, and Thien Nguyen. 2023.
\newblock Chatgpt beyond english: Towards a comprehensive evaluation of large language models in multilingual learning.
\newblock In \emph{Findings of the Association for Computational Linguistics: EMNLP 2023}, pages 13171--13189.

\bibitem[{Lester et~al.(2021)Lester, Al-Rfou, and Constant}]{lester2021power}
Brian Lester, Rami Al-Rfou, and Noah Constant. 2021.
\newblock The power of scale for parameter-efficient prompt tuning.
\newblock In \emph{Proceedings of the 2021 Conference on Empirical Methods in Natural Language Processing}, pages 3045--3059.

\bibitem[{Li and Liang(2021)}]{li2021prefix}
Xiang~Lisa Li and Percy Liang. 2021.
\newblock Prefix-tuning: Optimizing continuous prompts for generation.
\newblock In \emph{Proceedings of the 59th Annual Meeting of the Association for Computational Linguistics and the 11th International Joint Conference on Natural Language Processing (Volume 1: Long Papers)}, pages 4582--4597.

\bibitem[{Lialin et~al.(2023)Lialin, Deshpande, and Rumshisky}]{lialin2023scaling}
Vladislav Lialin, Vijeta Deshpande, and Anna Rumshisky. 2023.
\newblock Scaling down to scale up: A guide to parameter-efficient fine-tuning.
\newblock \emph{arXiv preprint arXiv:2303.15647}.

\bibitem[{Liu et~al.(2022)Liu, Ji, Fu, Tam, Du, Yang, and Tang}]{liu2022p}
Xiao Liu, Kaixuan Ji, Yicheng Fu, Weng Tam, Zhengxiao Du, Zhilin Yang, and Jie Tang. 2022.
\newblock P-tuning: Prompt tuning can be comparable to fine-tuning across scales and tasks.
\newblock In \emph{Proceedings of the 60th Annual Meeting of the Association for Computational Linguistics (Volume 2: Short Papers)}, pages 61--68.

\bibitem[{Liu et~al.(2024)Liu, Zheng, Du, Ding, Qian, Yang, and Tang}]{liu2024gpt}
Xiao Liu, Yanan Zheng, Zhengxiao Du, Ming Ding, Yujie Qian, Zhilin Yang, and Jie Tang. 2024.
\newblock Gpt understands, too.
\newblock \emph{AI Open}, 5:208--215.

\bibitem[{Mageed et~al.(2021)Mageed, Elmadany et~al.}]{mageed2021arbert}
Muhammad~Abdul Mageed, Abdelrahim Elmadany, et~al. 2021.
\newblock Arbert \& marbert: Deep bidirectional transformers for arabic.
\newblock In \emph{Proceedings of the 59th Annual Meeting of the Association for Computational Linguistics and the 11th International Joint Conference on Natural Language Processing (Volume 1: Long Papers)}, pages 7088--7105.

\bibitem[{Muhammad et~al.(2020)Muhammad, Chiyu, Houda, and Nizar}]{muhammad2020nadi}
Abdul-Mageed Muhammad, Zhang Chiyu, Bouamor Houda, and H~Nizar. 2020.
\newblock Nadi 2020: The first nuanced arabic dialect identification shared task.
\newblock In \emph{Proceedings of the Fifth Arabic Natural Language Processing Workshop}, pages 97--110.

\bibitem[{OpenAI(2023)}]{openai2023chatgpt}
OpenAI. 2023.
\newblock Chatgpt.
\newblock Version: GPT-4, Mar 14. Available at https://chat.openai.com.

\bibitem[{Pan(2020)}]{pan2020transfer}
Sinno~Jialin Pan. 2020.
\newblock Transfer learning.
\newblock \emph{Learning}, 21:1--2.

\bibitem[{Rane et~al.(2024)Rane, Choudhary, and Rane}]{rane2024gemini}
Nitin Rane, Saurabh Choudhary, and Jayesh Rane. 2024.
\newblock Gemini versus chatgpt: applications, performance, architecture, capabilities, and implementation.
\newblock \emph{Journal of Applied Artificial Intelligence}, 5(1):69--93.

\bibitem[{Sun et~al.(2023)Sun, Li, Li, Wu, Guo, Zhang, and Wang}]{sun2023text}
Xiaofei Sun, Xiaoya Li, Jiwei Li, Fei Wu, Shangwei Guo, Tianwei Zhang, and Guoyin Wang. 2023.
\newblock Text classification via large language models.
\newblock In \emph{Findings of the Association for Computational Linguistics: EMNLP 2023}, pages 8990--9005.

\bibitem[{Talafha et~al.(2020)Talafha, Ali, Za’ter, Seelawi, Tuffaha, Samir, Farhan, and Al-Natsheh}]{talafha2020multi}
Bashar Talafha, Mohammad Ali, Muhy~Eddin Za’ter, Haitham Seelawi, Ibraheem Tuffaha, Mostafa Samir, Wael Farhan, and Hussein Al-Natsheh. 2020.
\newblock Multi-dialect arabic bert for country-level dialect identification.
\newblock In \emph{Proceedings of the Fifth Arabic Natural Language Processing Workshop}, pages 111--118.

\bibitem[{Yang et~al.(2019)Yang, Dai, Yang, Carbonell, Salakhutdinov, and Le}]{yang2019xlnet}
Zhilin Yang, Zihang Dai, Yiming Yang, Jaime Carbonell, Russ~R Salakhutdinov, and Quoc~V Le. 2019.
\newblock Xlnet: Generalized autoregressive pretraining for language understanding.
\newblock \emph{Advances in neural information processing systems}, 32.

\bibitem[{Zaidan and Callison-Burch(2011)}]{zaidan2011arabic}
Omar Zaidan and Chris Callison-Burch. 2011.
\newblock The arabic online commentary dataset: an annotated dataset of informal arabic with high dialectal content.
\newblock In \emph{Proceedings of the 49th Annual Meeting of the Association for Computational Linguistics: Human Language Technologies}, pages 37--41.

\bibitem[{Zaidan and Callison-Burch(2014)}]{zaidan2014arabic}
Omar~F Zaidan and Chris Callison-Burch. 2014.
\newblock Arabic dialect identification.
\newblock \emph{Computational Linguistics}, 40(1):171--202.

\bibitem[{Zampieri et~al.(2017)Zampieri, Malmasi, Ljube{\v{s}}i{\'c}, Nakov, Ali, Tiedemann, Scherrer, and Aepli}]{zampieri2017findings}
Marcos Zampieri, Shervin Malmasi, Nikola Ljube{\v{s}}i{\'c}, Preslav Nakov, Ahmed Ali, J{\"o}rg Tiedemann, Yves Scherrer, and No{\"e}mi Aepli. 2017.
\newblock Findings of the vardial evaluation campaign 2017.
\newblock In \emph{Proceedings of the fourth workshop on NLP for similar languages, varieties and dialects}.

\bibitem[{Zampieri et~al.(2018)Zampieri, Malmasi, Nakov, Ali, Shon, Glass, Scherrer, Samard{\v{z}}i{\'c}, Ljube{\v{s}}i{\'c}, Tiedemann et~al.}]{zampieri2018language}
Marcos Zampieri, Shervin Malmasi, Preslav Nakov, Ahmed Ali, Suwon Shon, James Glass, Yves Scherrer, Tanja Samard{\v{z}}i{\'c}, Nikola Ljube{\v{s}}i{\'c}, J{\"o}rg Tiedemann, et~al. 2018.
\newblock Language identification and morphosyntactic tagging. the second vardial evaluation campaign.

\end{thebibliography}
\clearpage
\appendix
 \begin{figure*}
    \centering
    \includegraphics[width=1.0\linewidth]{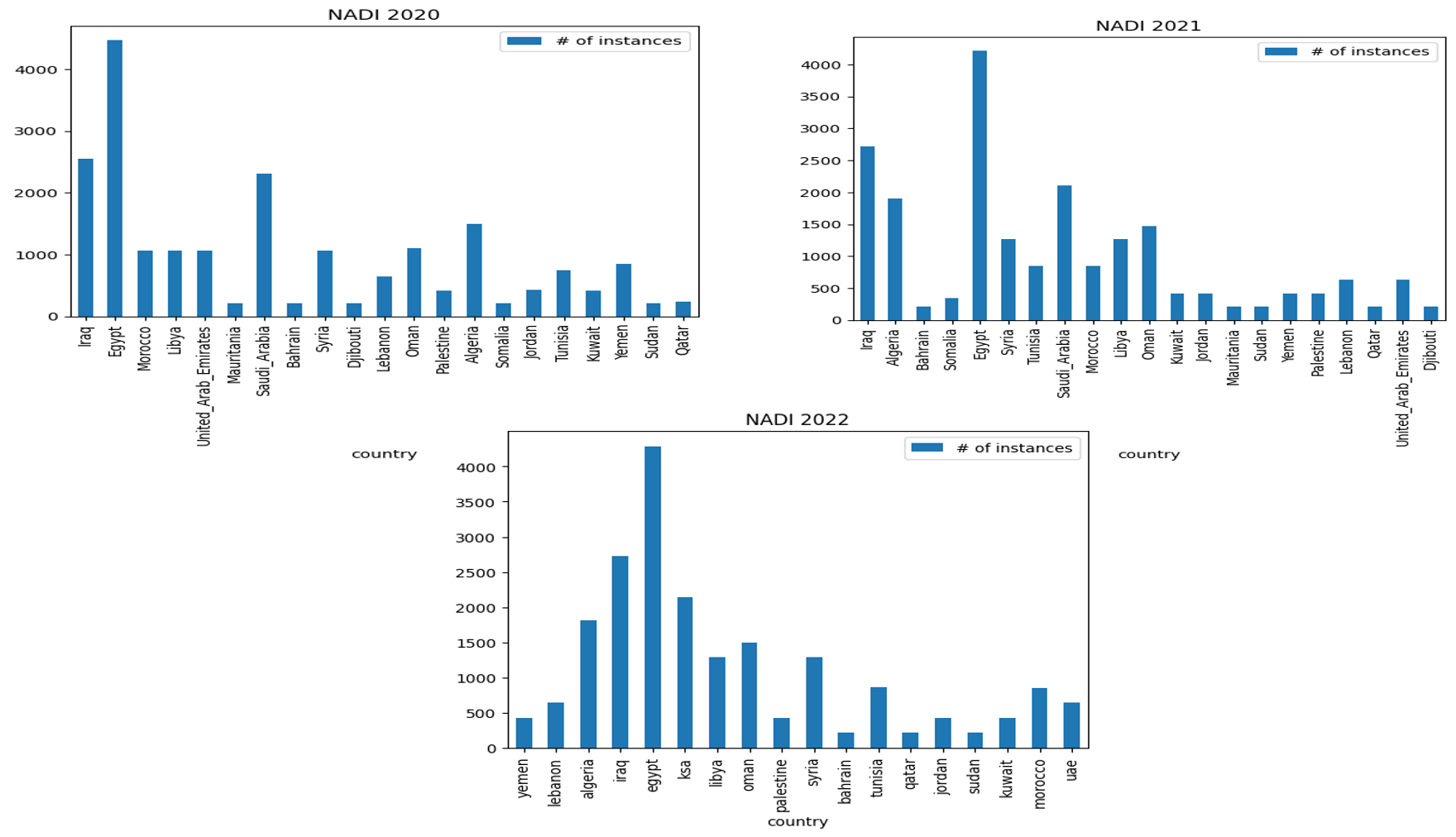}
    \caption{Label distributions of NADI 2020-2022}
    \label{fig:country}
\end{figure*}

\section{Dataset details} \label{app:data}
Figure \ref{fig:country} represents the label distributions of NADI 2020-2022. NADI-2023 has a balanced distribution.
\section{Prompt Details}\label{app:prompts}
The main prompt templates are given in Figures \ref{fig:prompt}, \ref{fig:prompt2}, and \ref{fig:prompt3}.
\floatstyle{boxed}
\restylefloat{figure}
\begin{figure*}[htp!]
\centering
\tiny
\begin{verbatim}
system_msg = You are a dialect classification model that is really good at following instructions. Please follow the user's instructions as precisely as you can.

user_prompt =  Your task will be to classify the given Arabic text into one of the following classes: {classes}.

Please respond with a single label that you think fits the text best. Each of the input text belongs to a specific Arabic dialect. 
Analyze the dialectal features and then give the prediction. Classify the following list 
of Arabic dialectal texts:
text: '{X}'
class:
\end{verbatim}
\caption{Vanilla zero-shot prompt.}\label{fig:prompt}
\end{figure*}

\floatstyle{boxed}
\restylefloat{figure}
\begin{figure*}[htp!]
\centering
\tiny
\begin{verbatim}
user_prompt = Your task will be to classify the given Arabic text and decide whether it belongs to a given dialect or not. 
The dialect classes are given: {classes}. Each of the input text belongs to a specific Arabic dialect. 
Analyze the dialectal features and then give your prediction. Let us follow a step by step process.

1. Assume yourself as a binary classifier
2. Ask whether the text belongs to each of the dialects in the given 
list or not: {classes}.
3. Output the predictions
Arabic Input text: '{X}'
predictions:
"""
\end{verbatim}
\caption{Binary zero-shot prompt.}\label{fig:prompt2}
\end{figure*}

 \floatstyle{boxed}
\restylefloat{figure}
\begin{figure*}[htp!]
\centering
\tiny
\begin{verbatim}
You are a dialect classifier. List the most important textual features of the Arabic dialects from the given list if you need to classify them. 
Please note that classification is based only on the written text. Can you provide a dialectal vocabulary of 20-30 words for each dialect?.  

The dialects are: 
["Algeria","Bahrain","Egypt","Iraq","Jordan","Kuwait","Lebanon","Libya","Morocco",
"Oman","Palestine","Qatar",
"Saudi_Arabia","Sudan","Syria","Tunisia","UAE","Yemen"]". 

Please give the answer in list of dictionaries with each dict key corresponding to 
the dialect and the values the vocab list. Include only the Arabic words.”
\end{verbatim}
\caption{Few-shot prompt inspired by CARP \cite{sun2023text}.}\label{fig:prompt3}
\end{figure*}

\end{document}